\documentclass{llncs}

\usepackage{makeidx}  

\usepackage{graphicx}
\usepackage{makeidx}  
\usepackage{subfigure}
\usepackage{color}
\usepackage{epstopdf}
\usepackage{cite}
\usepackage[linesnumbered,ruled,vlined]{algorithm2e}
\usepackage[english]{babel}
\usepackage{hhline}
\usepackage{amsmath,amssymb}
\usepackage{url}
\usepackage{tikz}

\begin{document}
\frontmatter          
\mainmatter              
\title{VOIDD: automatic vessel-of-intervention dynamic detection in PCI procedures}
\titlerunning{Vessel of intervention detection in PCI} 
\author{Ketan Bacchuwar\inst{1,2} \and Jean Cousty\inst{2}
R\'{e}gis Vaillant\inst{1} \and Laurent Najman\inst{2}}
\authorrunning{Bacchuwar \em{et. al.}}
\tocauthor{Ketan Bacchuwar, Jean Cousty, R\'{e}gis Vaillant, Laurent Najman}
\institute{General Electric Healthcare, Buc, France\\
\and
Universit\'{e} Paris-Est,
LIGM, A3SI, ESIEE Paris\\
\email{ketan.bacchuwar@ge.com}} 


\maketitle              

\begin{abstract} 
In this article, we present the work towards improving the overall workflow of the Percutaneous Coronary Interventions (PCI) procedures by capacitating the imaging instruments to precisely monitor the steps of the procedure. In the long term, such capabilities can be used to optimize the image acquisition to reduce the amount of dose or contrast media employed during the procedure. We present the automatic VOIDD algorithm to detect the vessel of intervention which is going to be treated during the procedure by combining information from the vessel image with contrast agent injection and images acquired during guidewire tip navigation. Due to the robust guidewire tip segmentation method, this algorithm is also able to automatically detect the sequence corresponding to guidewire navigation. We present an evaluation methodology which characterizes the correctness of the guide wire tip detection and correct identification of the vessel navigated during the procedure. On a dataset of 2213 images from 8 sequences of 4 patients, VOIDD identifies vessel-of-intervention with accuracy in the range of~$88\%$ or above and absence of tip with accuracy in range of~$98\%$ or above depending on the test case. 
\end{abstract}

\section{Introduction}
Percutaneous Coronary Intervention (PCI) is a procedure employed for the treatment of coronary artery stenosis. 
PCI is a very mature procedure relying on the deployment of a stent having the shape of the artery at the location of the stenosis.
These procedures are performed under X-ray guidance with use of contrast agent.
Consequently they also have side effects such as the injection of contrast agent based on iodine to the patient. 
The tolerance to this contrast agent is limited to some amount.
The other side effect is the use of ionizing radiation which affects both the patient and the operator.

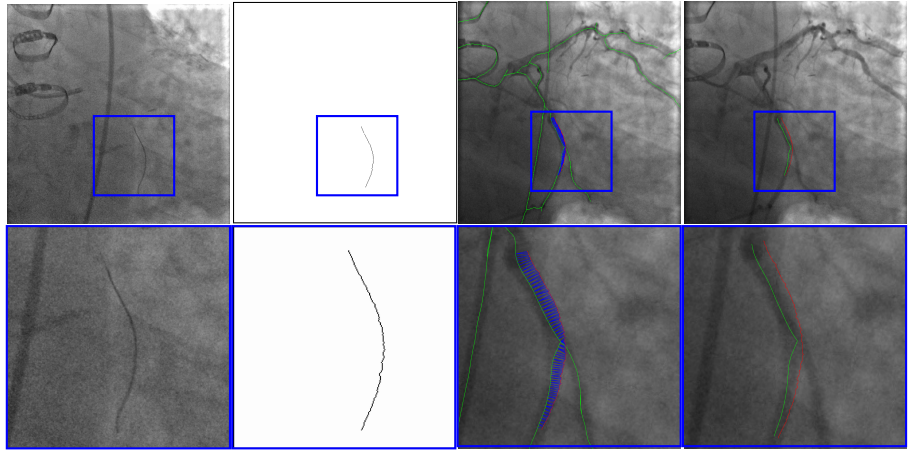
\begin{figure}[ht!]
\centering
\begin{tikzpicture}
\node[anchor=south west,inner sep=0](image) at (0,0) {\includegraphics[width=0.242\textwidth]{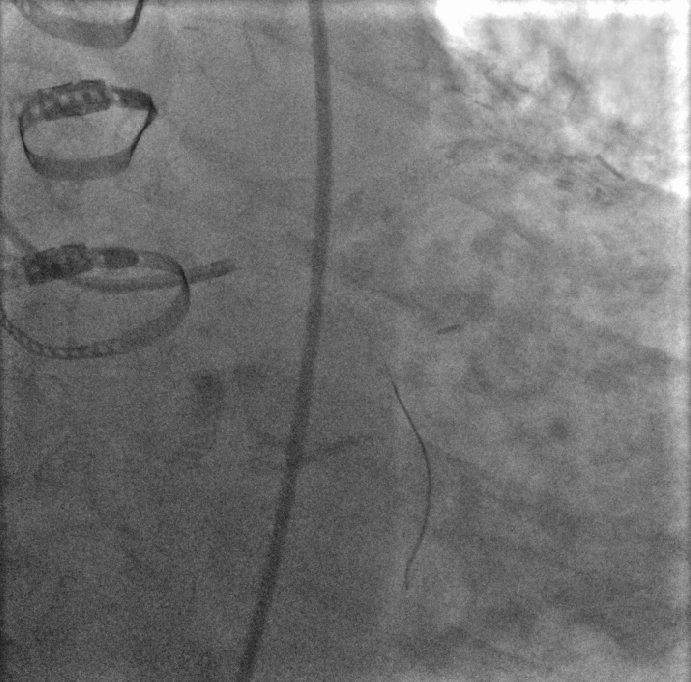}};
\node[anchor=south west,inner sep=0] at (3,0) {\setlength{\fboxsep}{0pt}%
    \fbox{\includegraphics[width=0.242\textwidth]{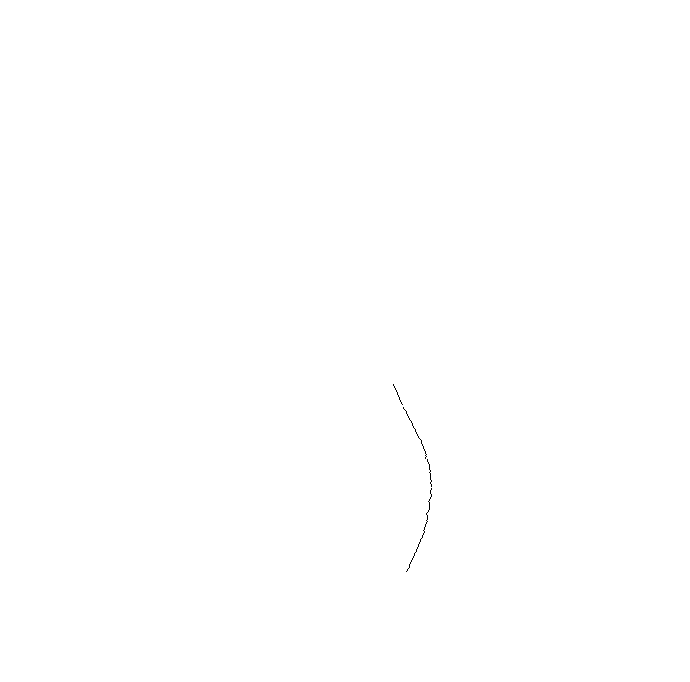}}};
\node[anchor=south west,inner sep=0] at (6,0) {\includegraphics[width=0.242\textwidth]{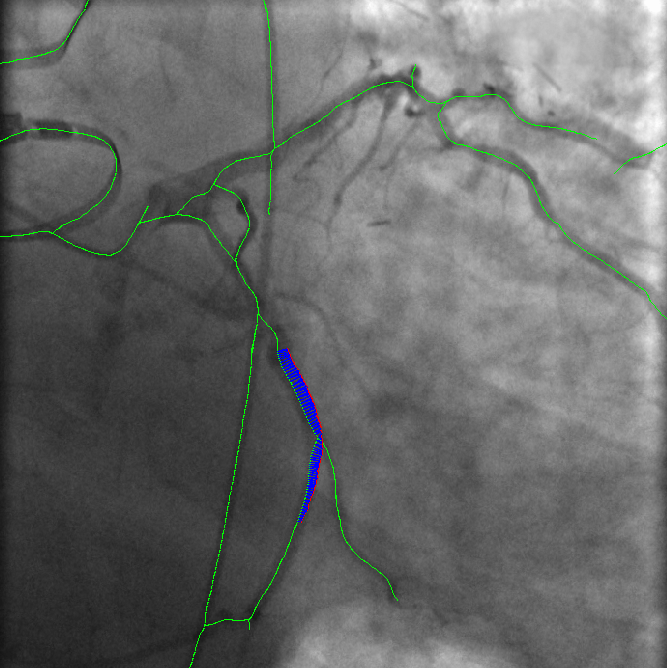}};
\node[anchor=south west,inner sep=0] at (9,0) {\includegraphics[width=0.242\textwidth]{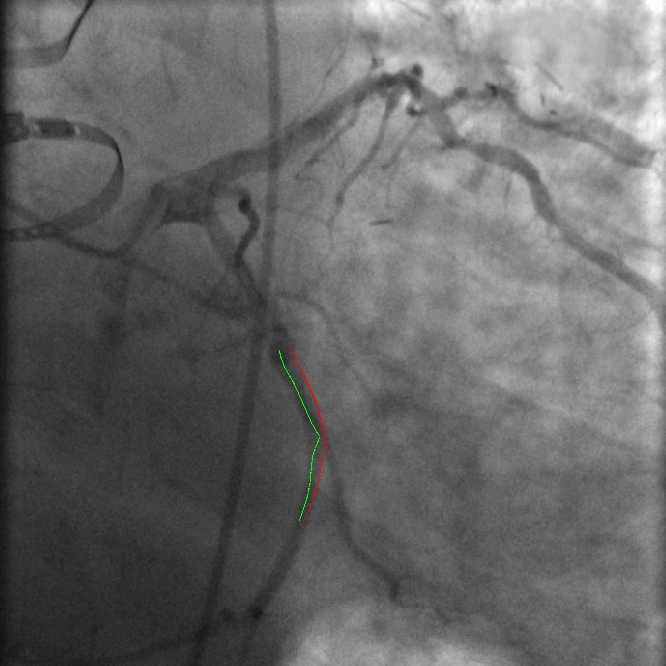}};
\node[anchor=south west,inner sep=0] at (0,-3) {\includegraphics[width=0.242\textwidth]{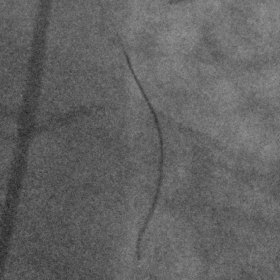}};
\node[anchor=south west,inner sep=0] at (3,-3) {\includegraphics[width=0.242\textwidth]{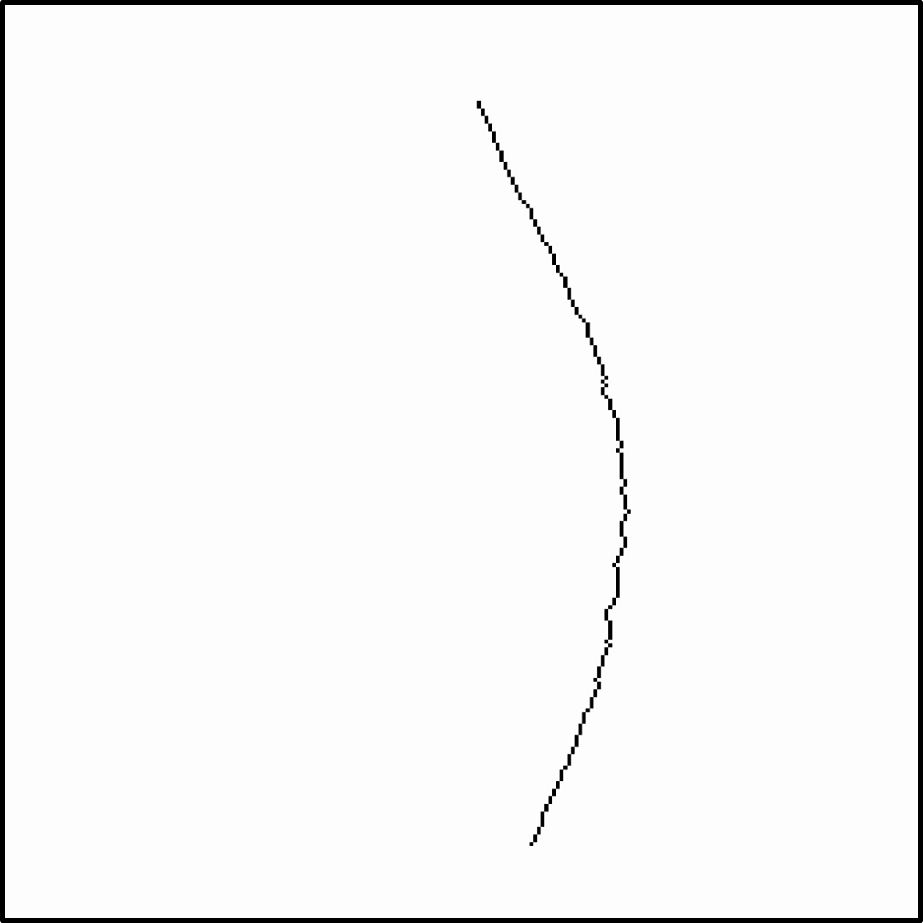}};
\node[anchor=south west,inner sep=0] at (6,-3) {\includegraphics[width=0.242\textwidth]{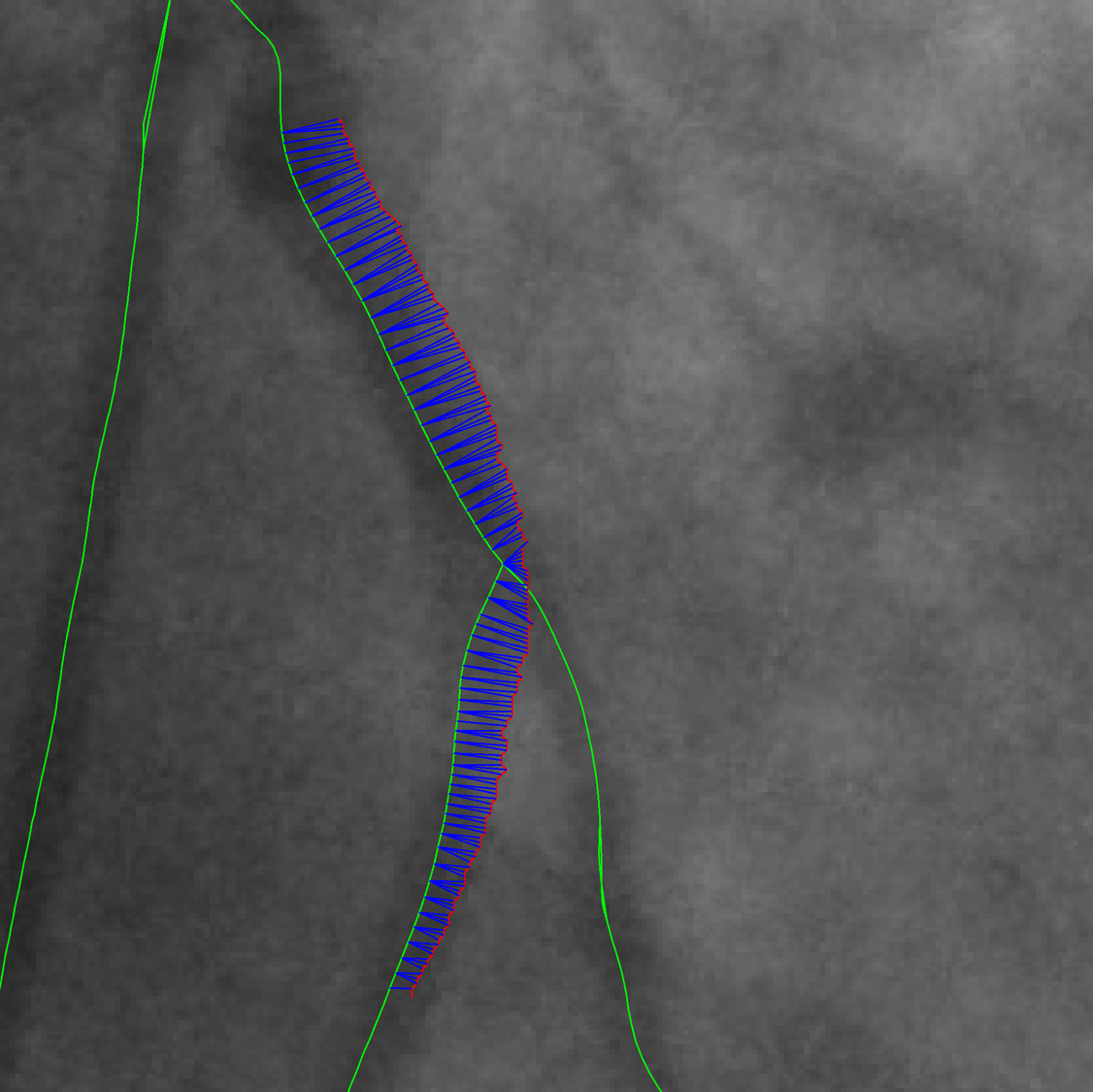}};
\node[anchor=south west,inner sep=0] at (9,-3) {\includegraphics[width=0.242\textwidth]{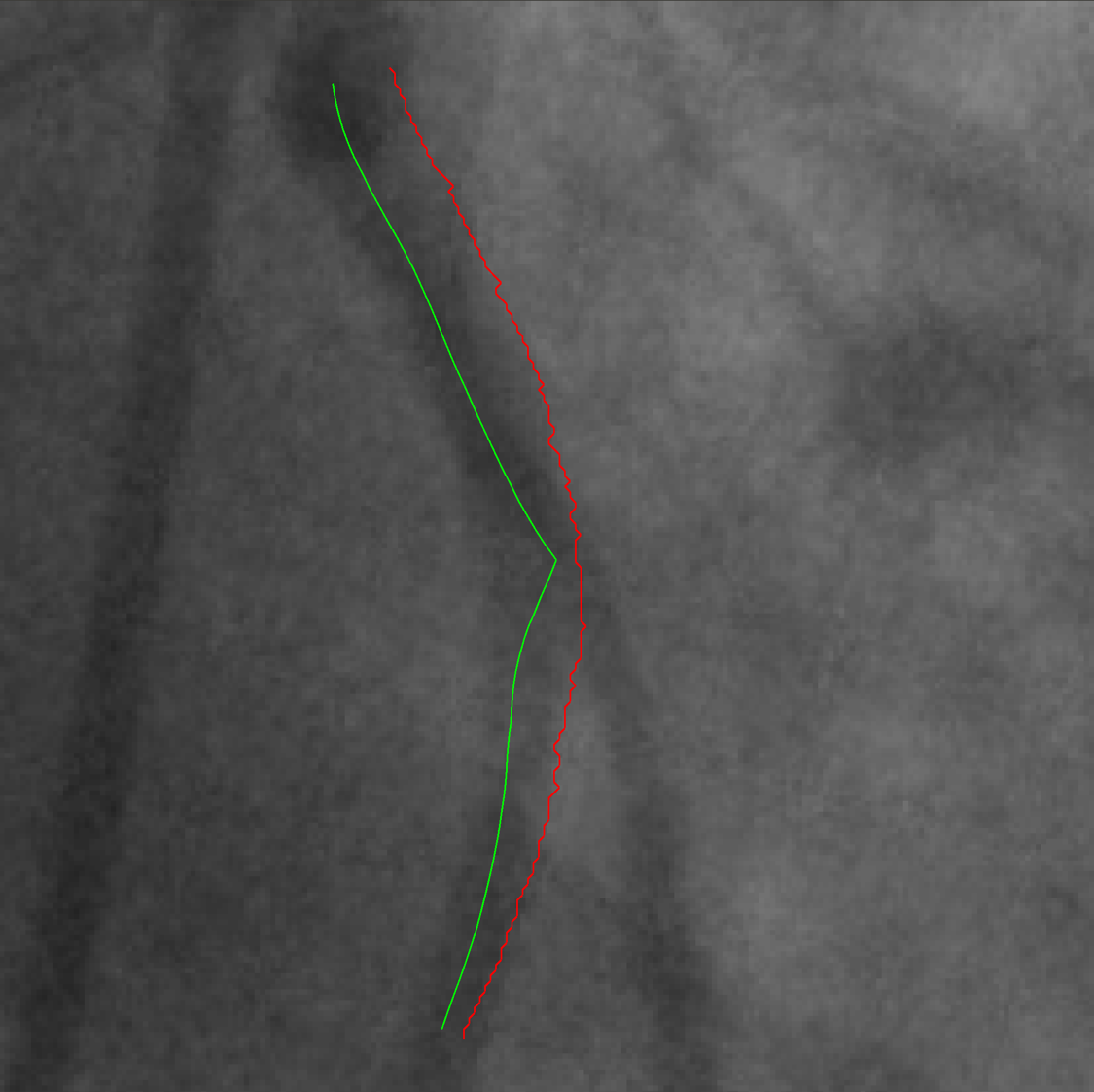}};

	\begin{scope}[x={(image.south east)},y={(image.north west)}]    
	    \draw[blue,thick] (0.39,0.13) rectangle (0.75,0.49);
	    \draw[blue,thick] (1.39,0.13) rectangle (1.75,0.49);
	    \draw[blue,thick] (2.35,0.15) rectangle (2.71,0.51);
	    \draw[blue,thick] (3.35,0.15) rectangle (3.71,0.51);
	    \draw[blue,thick] (0.0,-0.01) rectangle (1,-1.02);
	    \draw[blue,thick] (1.01,-0.01) rectangle (2.01,-1.02);
  	    \draw[blue,thick] (2.02,-0.01) rectangle (3.02,-1.01);
	    \draw[blue,thick] (3.03,-0.01) rectangle (4.03,-1.01);
	\end{scope}
\end{tikzpicture}
\caption{VOIDD: (from left to right)Input image~$f$; centerline of segmented guidewire tip; tip candidate (red) matched to vessel centerline (green) marked by pairings(blue); corresponding location (green) of guidewire tip(red) inside vessel.}
\label{fig:Method} 
\end{figure}

In the work presented here, we develop methods based on image processing to combine the information from fluoroscopic image sequences acquired at different steps of the procedure. 
More precisely, we consider two types of images: i) the images from {\em reference sequence}, which are injected with contrast agent to depict the vasculature and ii) the images from {\em navigation sequence}, which are acquired during the navigation of the tool and especially the guide wire, which is navigated from the ostia of the coronary artery down to the distal part after crossing the lesion. 
The imaging of the vessel with contrast agent provides information on the potential location of the stenosis.
The ECG of the patient is recorded along with the images. Standard algorithm as~\cite{hoffmann2015robust} can then be used to identify the subset of the images where the coronary images are well opacified with the contrast agent. In this subset, a reference sequence of about 10 to 15 images is then selected that covers a full cardiac cycle and includes best opacified images. 
The navigation sequence is obtained with a low dose acquisition mode called fluoroscopy. 
The guidewire, a very thin (wire-like) object of diameter~$0.014''$ has two sections. The distal section, called as the tip, is more important and is enough radio opaque to be seen with low dose X-ray mode.
Our aim is to automatically identify navigation sequence and determine the vessel-of-intervention which is going to be treated in the following steps of the PCI procedure, such as lesion reparation with angioplasty balloon, stenting, post-dilatation.

Several authors have worked on the task of segmenting the guidewire. 
For electrophysiology clinical application as in~\cite{milletari2014fully}, the size of the tip of the catheters makes its contrast significant enough to enable the development of robust algorithms. 
For PCI application as in~\cite{honnorat2012graph}, the weak contrast of the guidewire body makes the task very challenging. Some manufacturers of interventional suite have proposed or are still including in their offer, applications which facilitate the visual appreciation of the relationship between the guidewire and the vessel. The main idea is to combine a suite of consecutive injected images which visualize the vessel along a cardiac cycle. These images are combined with the images obtained during tool navigation. The images between these different times are paired mostly based on the ECG and up to our knowledge neither the breathing motion, nor any slight deformation of the arteries caused by the introduction of the guidewire are compensated. In~\cite{prasad2015co}, the correspondence between a location identified in the fluoroscopic images acquired during tool navigation and the cine images which depict the injected vessels is searched. The addressed clinical need is the registration of intra-vascular images acquired with a sensor placed along the guidewire with the vessel. By this means, the operator can easily correlate the readings of the angiographic images and the intravascular images/signals. In this situation, a full application is developed with a specific acquisition workflow with the different steps of the image acquisition and processing being done based on landmark points and appropriate images selected by an operator. 

The main contribution of this article is the proposition and the assessment of a method, called VOIDD, to automatically detect the so-called vessel-of-intervention during the navigation of the guidewire. 
More precisely this algorithm is able to recognize from the stream of fluoroscopic images following the acquisition of the reference sequence, the period corresponding to the guidewire navigation and to exploit it to determine the vessel-of-intervention (see Fig. \ref{fig:Method}). In order to reach this goal a general tracking algorithm is proposed and explained in section \ref{subsec:tracking}. This algorithm relies on features extracted from the navigation and reference images. Various methods can be adopted or designed to extract these features to be used with our general tracking algorithm. In this article, these features consists of vessel tree segmentation and of guidewire tip location candidates detection with advanced approaches involving the use of min tree~\cite{salembier2009connected}. Graph-based matching approaches derived from~\cite{benseghir2015tree} are used to match the guidewire tip with the vessel. These developments have been evaluated on 4 patient dataset. We present an evaluation methodology which characterizes the correctness of the guide wire tip detection and the correct identification of the vessel navigated during the procedures. On a dataset of 4 patients, VOIDD identifies vessel-of-intervention with accuracy in the range of~$88\%$ or above and absence of tip with accuracy in range of~$98\%$ or above depending on the test case.

\section{Vessel-of-intervention dynamic detection (VOIDD) algorithm}
\label{sec:method}
In this section, we first elaborate the general tracking framework of the VOIDD algorithm proposed in this article (in section \ref{subsec:tracking}). We then explain (in section \ref{subsec:pairedcandidates}) how to extract the features (from the {\em{reference sequence}} and the {\em{navigation sequence}}), which are used by the VOIDD algorithm. 

\begin{algorithm}
    \KwData{Guidewire navigation sequence and reference sequence~$\mathcal{R}$}
    \KwResult{$T_{vessel}$ Track of vessel-of-intervention and detected guidewire tips}
    \BlankLine
    Initialize $\mathcal{T}, T_{best} = \emptyset$ and $d_{best}$ to track assignment distance threshold \;
    \ForEach{image $I$ in the guidewire navigation sequence}
    {
        $\mathcal{P} := $ \textbf{ExtractFeaturePairs}(I,$\mathcal{R}$)\; 
        \tcp{feature pairs are ranked in decreasing order of matching score}
        \ForEach {~$P \in \mathcal{P}$}
        {
           \ForEach {~$T \in \mathcal{T}$}
           { 
               $d_{ij} := T\rightarrow$ \textbf{TrackAssignmentDistance}($P$) \;
               \If {~$d_{ij} < d_{best}$}
               {
                   $T_{best} := T$ ;
                   $d_{best} := d_{ij}$\;
               }
           }
           \If{$T_{best} \neq \emptyset $}
           {\textbf{AssignTrack}($T_{best}$,$P$)\;}
           \If{($P\rightarrow$ TrackNotAssigned())}
           {
               $T_{new}:=$ \textbf{MakeTrack($P$)};
               $\mathcal{T} \rightarrow$ AddTrack($T_{new}$) \;
           }
           Reset($T_{best},d_{best}$)\;
        }
    }
    $T_{vessel}$ = $\mathcal{T} \rightarrow $ LongestTrack() \;
    \caption{VOIDD}    
    \label{algorithm:voidd}
\end{algorithm}

\subsection{General tracking framework}
\label{subsec:tracking}
We aim to obtain the vessel-of-intervention by making a smart correspondence between the input guidewire navigation sequence and the reference sequence. 
Therefore, we propose an algorithm, called VOIDD, that is able to simultaneously detect the guidewire tip in the navigation sequence and the section of the coronary artery tree in which the guidewire is currently navigating in the reference sequence.
From a broader perspective, the algorithm consists of: i) detecting feature pairs from the navigation and reference sequence; ii) grouping these feature pairs into tracks, a track being a sequence of features that are spatially consistent in time; iii) selecting the most relevant track as the detected vessel-of-intervention. 
A feature pair is made of two corresponding curves. The first one, called a {\em tip candidate}, is extracted from the guidewire navigation sequence and possibly corresponds to the guidewire tip in the fluoroscopic image. The second one, called a vessel-of-intervention {\em (VOI) candidate} is obtained from the reference sequence and is a part of the coronary vessels that optimally fits the associated tip candidate. The precise description of the VOIDD algorithm and of the feature pairs extraction is given in Algorithm 1 and in the section \ref{subsec:pairedcandidates} respectively. 

VOIDD algorithm manages a dictionary of tracks~$\mathcal{T}$, where each track~$T \in \mathcal{T}$ is a sequence of feature pairs, with at most one pair per image in the guidewire navigation sequence.
For each time step of the navigation sequence, the essence of the algorithm lies in optimally assigning the best detected feature pair to the existing tracks. 
To this end, the feature pairs are ranked in decreasing order of matching score, provided by the feature extraction algorithm.
Then, a distance between feature pair and track, called the {\em track assignment distance} (TAD) (described in the section~\ref{subsec:trackassignment}), is considered to optimally assign the considered feature pair to the track which is at least distance.  
A TAD threshold is computed as the theoretically maximum possible value of TAD based on the length of the guidewire tip and maximum observed guidewire speed. If TAD to the closest track is above the TAD threshold then the feature pair is not assigned to any existing track but is used to initialize a new track in~$\mathcal{T}$. Once all the frames in the navigation sequence are processed, the longest track ({\em i.e.} track with maximum feature pairs) is selected as the vessel-of-intervention.  

\subsection{Feature pairs extraction}
\label{subsec:pairedcandidates}
This section elaborates the extraction of the feature pairs, which are associations between the images of the navigation sequence and the images of the reference sequence.
First, we explain the {\em tip candidate} extraction by segmentation and morphological thinning. 
This is followed by the extraction of centerline of the injected vessels to obtain vessel graph.
Finally, we present the matching part to find the possible associations (the {\em VOI candidates}) of the tip candidate in the vessel graph. However, different methods can be adopted or designed to obtain these feature pairs.

\textbf{Tip candidate extraction.} Guidewire tip appears as contrasted thin and elongated object in the fluoroscopic image.
We are interested to segment the guidewire tip, using a component tree called min tree.
The min tree \cite{salembier2009connected} structures all the connected components of the lower-level sets of the grayscale image based on inclusion relationship.
We assign to any connected component~$C$ of the min tree~$\mathcal{M}$, a shape attribute characterizing the shape and structural properties of guidewire tip. 
Then, the considered attribute~$\mathcal{A}$ describes the elongation of the components. For any component~$C$,~$\mathcal{A}(C) = {(\pi \times {l_{max}(C)}^2)}/{|C|} \enspace ,$ where~$|.|$ represents cardinality and~$l_{max}(C)$ is the length of the largest axis of the best fitting ellipse for the connected component~$C$. 
Since the guidewire tip is thin and long, the component corresponding to the tip have high value of attribute~$\mathcal{A}$. 
A mere thresholding of the elongation~$\mathcal{A}$ is not sufficient, often giving other long and elongated (unwanted) objects like pacing lead and filled catheters. 
Indeed, these objects have higher elongation value than the guidewire tip.
Hence, according to physical properties of the guidewire tip, we set a upper bound value~$t_{max}$ on~$\mathcal{A}$ to maximum possible elongation value of the guidewire tip, to ensure that extracted components contain guidewire tip. 
Even with this upperbound threshold keeping the most elongated component does not always lead to the desired tip. Based on min tree structure, the nested connected components that satisfy the criterion are filtered to preserve the component with largest area (taking aid of the inclusion relationship).
Therefore, we adopt the shaping framework~\cite{xu2012morphological} that allows us to efficiently extract significant connected components. The extracted components constitute the tip candidates. 
Shaping extensively uses the min tree structure to regularize the attributes and to select the relevant components. 
In order to facilitate matching, we perform skeletonization~\cite{coupriehal00727377} of the selected connected component(s) to obtain centerline of the tip candidates. 
Fig.~\ref{fig:Method} shows the obtained centerline of segmented guidewire tip from the input image.  This centerline of the tip candidate~$\mathcal{C}$ is modeled as a discrete polygonal curve.   

\textbf{Vessel centerline extraction.} The coronary vessels from each fluoroscopic image in the reference sequence are enhanced using a Hessian based technique~\cite{krissian2000model} followed by centerline extraction using Non-Maximum Supression and hysteresis thresholding. 
We represent these centerlines of vessels by a non-directed graph~$\mathcal{X}$ where the nodes are represented by bifurcations whereas the edges refer to curvilinear centerlines.
Apparent bifurcations resulting from superimposing vessels in 2D X-ray projections also form nodes. Such graph is computed for each frame in the reference sequence providing us with a representation for each phase of the cardiac cycle. 

\textbf{Matching.} An important step in the task of vessel-of-intervention detection is to designate possible desirable associations of the corresponding location of the guidewire tip inside the injected vessel. We refer to these locations in vessels as {\em vessel-of-interest }(VOI) candidates.  
This step refers to building the correspondences between the centerline of each tip candidate~$\mathcal{C}$ extracted from navigation sequence and the corresponding centerlines of the vessels~$\mathcal{X}$ extracted from reference sequence by taking into account ECG information. 
We adopt the curve pairing algorithm of~\cite{benseghir2015tree} to perform this task.
It is required to define a curve-to-curve distance to compare the two sets of curves mentioned above. 
We use a discrete version of Fr\'{e}chet distance~\cite{alt1992measuring} as it takes into account the topological structures of the curves. 
Thus, this Fr\'{e}chet distance is computed from a mapping between two ordered sets of discrete polygonal curves denoted by~$\mathcal{C}$ and by~$X_C$, respectively.
Imposed non-decreasing surjective mappings (reparameterization mapping) in computation of Fr\'{e}chet distance takes into account the order of points along curves.  
This order also helps us in curve pairing described below, to give scan direction along the curves. 

The above step requires the selection of every admissible curve~$X_C$ in graph~$\mathcal{X}$.
A curve in~$\mathcal{X}$ is a path between two nodes, without visiting the same edge twice. 
In order to restrict computational complexity of search, we restrict the set of admissible curves to be in the neighborhood of the tip candidate extremities $\mathcal{C}[1]$ and~$\mathcal{C}[n]$ and we construct all possible paths between them in the graph. 
Indeed, these admissible curves are the VOI candidates. These VOI candidates, together with the tip candidate~$\mathcal{C}$, is a{\em~set of feature pairs}~$\mathcal{P} = \{ (\mathcal{C},X_{C}) \mid X_{C}$ is some curve in $\mathcal{X}$ matched to $\mathcal{C}\} $, which are further filtered and ranked according to the shape similarity measure to prefer VOI candidates with higher shape resemblance to the tip candidate.
This term is computed from residual Fr\'{e}chet distance after the 2D transformation~\cite{benseghir2015tree}. 
The set of feature pairs~$\mathcal{P}$ is computed for each image in the guidewire tip navigation sequence by performing the matching with the vessel graph of the corresponding cardiac phase. 

\subsection{Track assignment distance (TAD)}
\label{subsec:trackassignment}
The TAD is computed as a distance between a proposed feature pair~$P = (\mathcal{C},X_{C})$ and a track~$T$. 
It is the average of tip candidate distance, VOI candidate distance and graph distance.
The tip candidate distance and the VOI candidate distance are computed between the proposed feature pair and the latest added feature pair of~$T$.
The tip candidate distance accounts for the geometrical shift between the two tip candidates. 
The VOI candidate distance measures the mean Euclidean distance between the end-points of the VOI candidates.
The graph distance is computed between the proposed feature pair and the latest iso cardiac phase feature pair in~$T$. 
It is the length of the path between two VOI candidates from two images in the same cardiac phase obtained from different cardiac cycles.
The VOI candidate distance and the graph distance helps to preserve temporal coherency in the tracks.
We transform these three distances with exponential functions so that they belong in the same range~$[0-1[$. The parameters of these exponential functions are set according to the length of the guidewire tip. 

\section{Results}
This section reports the performance of the VOIDD algorithm to detect the vessel-of-intervention and assesses its potential to identify the guidewire tip navigation sequence. 
An expert user annotated (with cross-validation) the centerline of the branch of the artery navigated by the guidewire tip as the ground truth. Ground truth was marked by a single expert user using a semi-automatic software guided by the vessel centerline extracted by the method in section \ref{subsec:pairedcandidates}.
This ground truth centerline is modeled as a discrete polygonal curve~$GT$ $=[GT[1]\cdots GT[M]]$.
A VOI candidate~$X$ selected by VOIDD is similarly modeled as~$X=[X[1]\cdots X[N]]$ with N equidistantly spaced points chosen at sub-pixel resolution. 
To assess the correctness of the automatically detected vessel, we consider the following target-to-registration (TRE) error between~$X$ and~$GT$ given by,~$TRE = \frac{1} {N} {\sum_{i=1}^{i=N}} \min\limits_{\forall j \in 1 \cdots M-1} \left| \mbox{d}(X_C[i],GT(j,j+1)) \right| ,$
where~$GT(j,j+1)$ refers to the segment between point~$GT[j]$ and~$GT[j+1]$ and d refers to point to segment distance converted to mm using known detector pixel size. 
If the tip is correctly paired to vessel-of-intervention then this TRE error is governed by the usual small difference between the estimated centerline and the expert marked vessel centerline.
The algorithm chosen tip and vessel-of-intervention are considered as a {\em correct detection} if the corresponding TRE error is less than~$0.5$mm.
If TRE error is more than~$0.5$mm, we consider that we have a {\em wrong detection}.
If the input image contains guidewire tip, but the algorithm do not provide any detection, then the TRE error cannot be computed and a {\em missed detection} is reported.
This may occur due to the fact that, sometimes the tip appears to be very blurred due to its sudden movement or due to reduced visibility of the tip caused by small contrast media injection to guide the navigation.  
In order to further evaluate the efficacy of the algorithm to identify the navigation sequence, we analyze its robustness to detect navigated vessel in sequence with no guidewire tip.
In such sequence, if the algorithm detects a vessel-of-intervention in an image, it is counted as a {\em false detection}.

\begin{table}[t]
\centering
\caption{Performance of VOIDD algorithm on 4 patients}
\label{table:1}
\begin{tabular}{|c|c|c|c|c|c|c|c|}
\hline
Patient & Sequence & \begin{tabular}[c]{@{}c@{}}Number \\ of \\ frames\end{tabular} & \begin{tabular}[c]{@{}c@{}}Frames \\ with\\ tips\end{tabular} & \begin{tabular}[c]{@{}c@{}}Correct\\ detection\end{tabular} & \begin{tabular}[c]{@{}c@{}}Wrong \\ detection\end{tabular} & \begin{tabular}[c]{@{}c@{}}Missed\\ detection\end{tabular} & \begin{tabular}[c]{@{}c@{}}False\\ Detection\end{tabular} \\ \hline
A       & A1       & 164                                                            & 164                                                           & 92.07\%                                                     & 0\%                                                        & 7.92\%                                                     & NA                                                        \\ \hline
B       & B1       & 706                                                            & 706                                                           & 88.52\%                                                     & 5.80\%                                                     & 5.66\%                                                     & NA                                                        \\ \hline
C       & C1       & 449                                                            & 449                                                           & 92.20\%                                                     & 5.12\%                                                     & 2.67\%                                                     & NA                                                        \\ \hline
D       & D1       & 204                                                            & 204                                                           & 89.70\%                                                     & 2.94\%                                                     & 7.35\%                                                     & NA                                                        \\ \hhline{|=|=|=|=|=|=|=|=|}
A       & A2       & 156                                                            & 0                                                             & NA                                                          & NA                                                         & NA                                                         & 1.28\%                                                    \\ \hline
B       & B2       & 172                                                            & 0                                                             & NA                                                          & NA                                                         & NA                                                         & 0.58\%                                                    \\ \hline
C       & C2       & 264                                                            & 0                                                             & NA                                                          & NA                                                         & NA                                                         & 1.50\%                                                    \\ \hline
D       & D2       & 98                                                             & 0                                                             & NA                                                          & NA                                                         & NA                                                         & 2.04\%                                                    \\ \hline
\end{tabular}
\end{table}

\begin{figure}[t]
\centering
\begin{tikzpicture}
\node[anchor=south west,inner sep=0](image) at (0,0) {\includegraphics[width=0.484\textwidth]{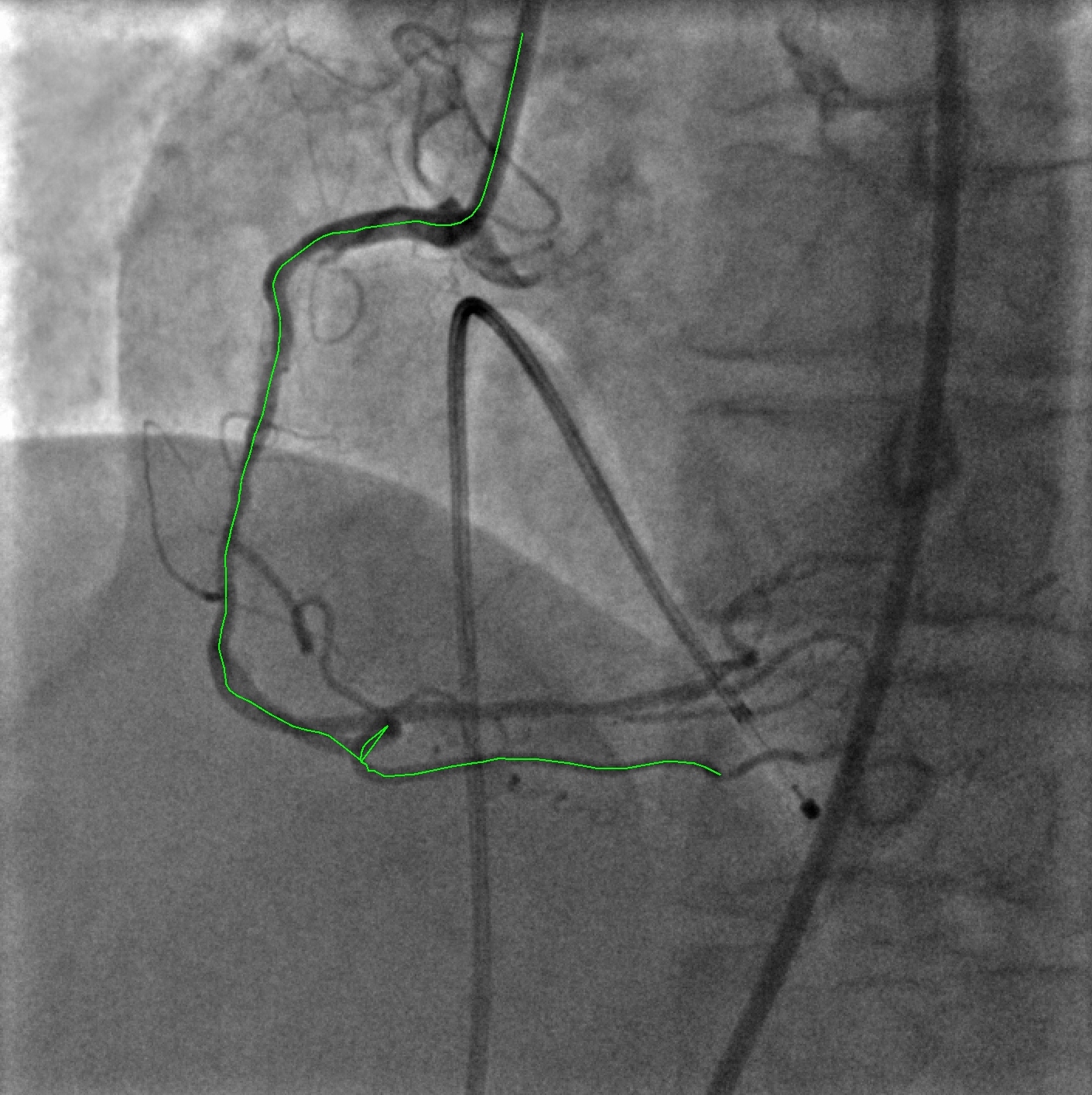}};
\node[anchor=south west,inner sep=0] at (6,0) {\includegraphics[width=0.484\textwidth]{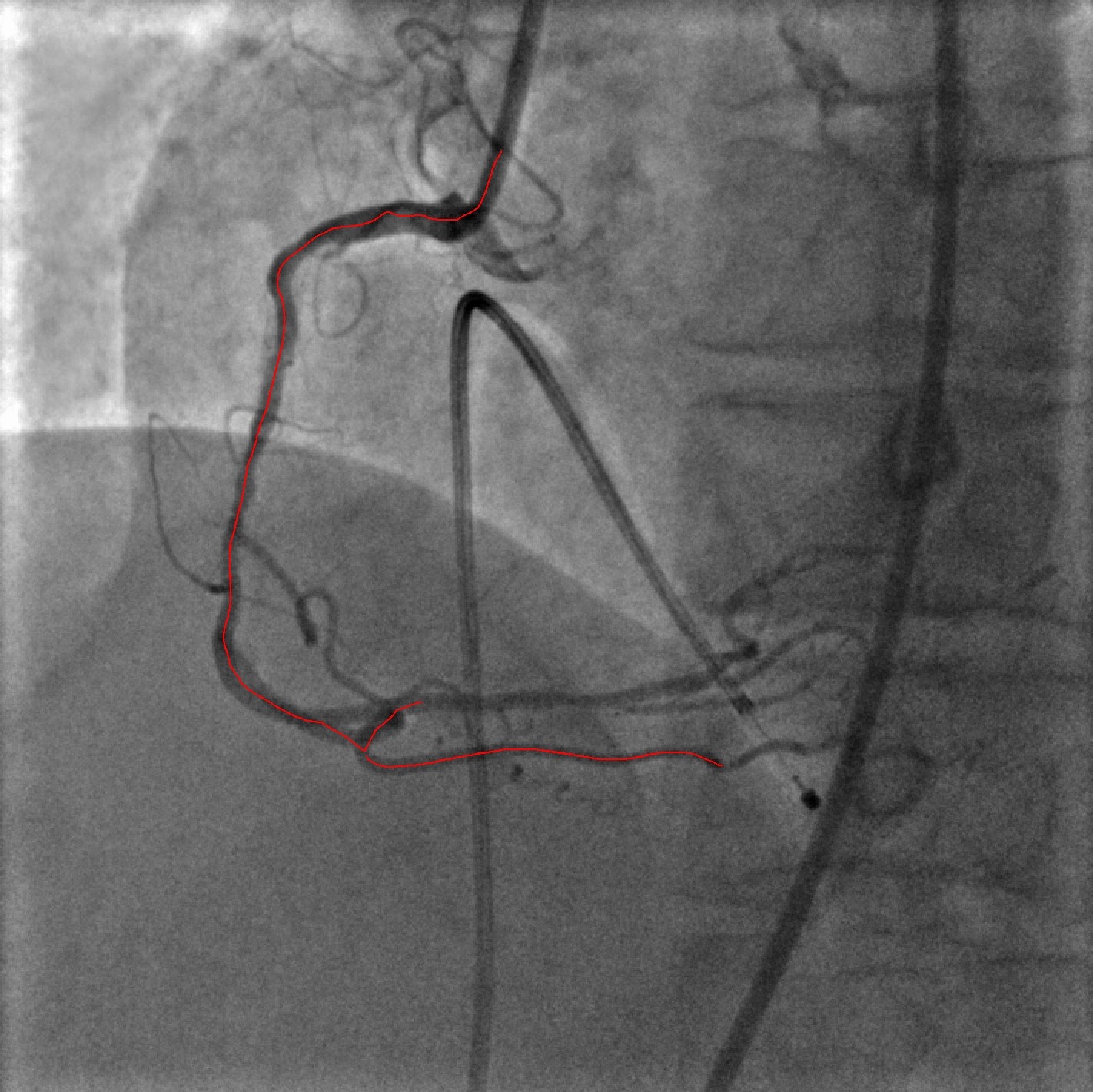}};
\draw [->,>=stealth, blue, very thick] (1.65,1.35) -- (1.95,1.78);
\end{tikzpicture}
\caption{VOIDD result: ground truth(in green) and vessel of intervention obtained from longest track(in red). In ground truth, the catheter is marked by the expert but not part of the tracked vessel because the guidewire tip is not detected when it is still inside the catheter. In this case, the vessel makes a very tight loop (blue arrow) in the bottom. In the tracking, we fail to detect this loop.}
\label{fig:results}  
\end{figure}

Sequences A1, B1, C1 and D1 in Table~\ref{table:1} show the efficacy of VOIDD algorithm to detect vessel-of-intervention during guidewire navigation in 4 patients and over 1513 images. In summary, VOIDD algorithm is able to correctly determine the location of tip in the vessel-of-intervention with an accuracy of around~$88\%-92\%$. 
The sequences A2, B2, C2 and D2 in the Table~\ref{table:1} portray the efficiency to identify navigation sequence over 690 images when guidewire tip is absent in the fluoroscopic images. The VOIDD algorithm is able to detect these sequences as sequence without guidewire tip with accuracy of~$98\%-99\%$. Analyzing the navigation sequence detection accuracy of VOIDD, we can use it to automatically detect the arrival of the guidewire tip. 
The parameters involved in various stages of the algorithm{\em e.g.} tip candidate extraction or TAD were designed based on the physical properties of guidewire tip, permissible speed of advancement of guidewire. Current implementation runs in average~$0.33$ seconds for tracking on a Intel Core i7 cadenced at~$2.80$ GHz. Videos are available as supplementary material\footnote{\url{https://voidd-miccai17.github.io/}}. Figure \ref{fig:results} shows the vessel of intervention obtained by the VOIDD algorithm and corresponding ground truth on left.

\section{Conclusion and future work}
We proposed in this paper a framework to determine the vessel-of-intervention in fluoroscopic images during the PCI procedures.
We also demonstrate the segmentation of the guidewire tip and the accuracy of its detection. 
This algorithm has the potential to be part of the software embarked by X-ray imaging systems and capable of automatically monitoring the successive steps of the procedure in view of continuously adapting the system behavior to the user needs.
For instance, the guidewire tip tracking can be used to determine the phases related to the navigation of the guidewire, adding more semantic information, hence can be a first step towards smart semantic monitoring of the procedure. In order to perform such semantic analysis of the procedure, it is important to know the position of different interventional tools like guidewire tip, marker balls, balloon and this application opens the doors to ease the segmentation of these objects in the vessel-of-intervention. 
Encouraging results have been obtained with success rate above~$88\%$ for vessel of intervention detection. Future work includes the collection of additional clinical cases. In the longer term, we will investigate the detection of the other major tools and their integration into a semantic model of the procedure.

\bibliographystyle{splncs03}      
\bibliography{llncs} 
\end{document}